\documentclass[pmlr]{jmlr}

\usepackage{longtable}

\usepackage{mathabx}

 \usepackage{booktabs}
 
\usepackage[load-configurations=version-1]{siunitx} 

\makeatletter
\def\set@curr@file#1{\def\@curr@file{#1}} 
\makeatother

\theorembodyfont{\upshape}
\theoremheaderfont{\scshape}
\theorempostheader{:}
\theoremsep{\newline}

\jmlrvolume{}
\jmlryear{2023}
\jmlrworkshop{Machine Learning for Healthcare}

\title[Data augmentation method for modeling health records]{Data augmentation method for modeling health records \\
with applications to clopidogrel treatment failure detection}

\author{\Name{Sunwoong Choi}
       \Email{tony.choi@cipherome.com}\\ 
       \addr Cipherome Korea\\       
       Seoul, Korea
       \AND
       \Name{Samuel Kim}
       \Email{samuel.kim@cipherome.com}\\ 
       \addr Cipherome Inc.\\
       San Jose, CA, U.S.A.} 

\editor{Editor's name}

\begin{document}

\maketitle
\begin{abstract}
We present a novel data augmentation method to address the challenge of data scarcity in modeling longitudinal patterns in Electronic Health Records (EHR) of patients using natural language processing (NLP) algorithms. The proposed method generates augmented data by rearranging the orders of medical records within a visit where the order of elements are not obvious, if any. Applying the proposed method to the clopidogrel treatment failure detection task enabled up to 5.3\% absolute improvement in terms of ROC-AUC (from 0.908 without augmentation to 0.961 with augmentation) when it was used during the pre-training procedure. It was also shown that the augmentation helped to improve performance during fine-tuning procedures, especially when the amount of labeled training data is limited.
\end{abstract}

\section{Introduction}
\label{sec:intro}
The Electronic Health Record (EHR) includes diagnoses, procedures, prescriptions, and other items of protected health information. This data is usually timestamped, allowing for researchers to study patients' longitudinal patterns. Various machine learning algorithms have been devised to model the longitudinal patterns using health records in EHR. For example, \cite{choi2017retain} proposed RETAIN  to detect heart failures using a recurrent neural network (RNN) architecture. 

Recently, researchers have applied natural language processing (NLP) algorithms to EHR data by drawing analogies between medical records and natural language. In particular, Bidirectional Encoder Representations from Transformers (BERT) algorithm~\cite{devlinBERTPretrainingDeep2019} has been widely used. It is a transformer-based algorithm that utilizes pre-trained models trained on a large unlabeled dataset and fine-tunes models with a labeled but relatively small dataset. \cite{shangPretrainingGraphAugmented2019} was for medication recommendation using a BERT-based graph neural network (GNN) model. \cite{li2019behrt} proposed BEHRT to learn patterns of diagnostic codes and used the model to predict unseen diagnoses. Similarly, \cite{rasmyMedBERTPretrainedContextualized2021} proposed Med-BERT which also adapted the BERT framework to predict diseases. \cite{pang2021cehrbert} incorporated temporal information using a hybrid approach by augmenting the input to BERT using artificial time tokens, incorporating time, age, and concept embedding, and introduced a new second learning objective for visit type. 
The researchers hypothesized that a patient's medical history could be structured like a natural language document, where patient's visit corresponded to sentences and the diagnoses, procedures, and medications correspond to words.

However, there are still challenges in applying NLP algorithms to build models using EHR data. One challenge is the relatively small amount of training data available for both pre-training and fine-tuning, compared to the amount of data available for NLP applications. Another challenge is that the analogy between natural language and medical history is not perfect; for example, forcing a certain order for medical records to form a sentence-like sequence to be fed into NLP algorithms may cause ordinal biases since the order of medical records during a single visit are rarely significant.

To address these challenges, we developed a data augmentation method to increase the amount of available training data. This method generates augmented data by rearranging the order of medical records within a visit. We hypothesize that this method will mitigate the challenges mentioned above by not only providing with more data but also mitigating ordinal biases. We argue that this study demonstrates the potential of applying NLP algorithms to EHR data, while also highlighting the challenges that need to be addressed.

We applied the proposed strategies to model drug responses with given EHR records. Drug responses, i.e., adverse drug reactions (ADRs) and treatment failures (TFs), can occur due to inadequate concentrations of medications. They significantly impact patients’ lives and are serious burdens to the healthcare system. As stated in \cite{Pitts2022}, ADRs are estimated to be between the fourth and sixth most common cause of death worldwide, taking their place among other prevalent causes of mortality such as heart disease, cancer, and strok, while TFs may occur when therapeutic doses are not reached. The variable presentation makes identifying drug responses difficult, and the impact is likely underrepresented~\cite{ADRUnderReporting2006}. The multi-factorial nature of drug responses adds to the complexity; patient’s age, renal or liver function, comorbidities, comedications, lifestyle, and genetic predispositions are all important factors affecting drug responses~\cite{SignalADR2008,PharmacogeneticsADR2000}.

Among the multitude of drugs, we studied clopidogrel, a P2Y12 inhibitor, one of the most widely prescribed drugs in the US. Its use with aspirin is commonly referred to as dual antiplatelet therapy (DAPT), and is the standard of care following stent placement for cardiovascular disease. However, various responses to this drug have been reported in several studies raising concerns for its safety~\cite{ResistanceAntiPLT2007}. High concentration of active metabolite can cause bleeding, while low concentrations of active metabolites can cause recurrent thrombosis, both of which can lead to medical emergencies. There is a clear need for a screening tool to better predict which patients may be at risk for such events.

\section{Generalizable insights}
We believed that our research underscores the need for innovative approaches to tackle the challenges of utilizing language models in healthcare applications where data availability may be limited. By addressing these challenges, our research contributes to more accurate and effective prediction of medical events, leading to improved patient care and outcomes in the healthcare domain.

\section{Data}
\label{sec:data}
\subsection{Dataset}
We used the UK Biobank which consists of data collected from 502,527 participants ~\cite{UKB2015}. Volunteers aged 40 to 70 were recruited from England, Scotland, and Wales and invited to assessment centers between 2006 and 2010. Data collected during the visits included biosamples, physical examination measurements, and questionnaire answers followed by interviews. Genomic data, both sequencing and genotyping, were generated using the biosamples collected.  Also, an extensive and comprehensive medical history data was made available through hospital in-patient and primary care data from external data sources. 

We extracted prescriptions, diagnoses and procedure records, along with dates, for all participants from the UK Biobank. All prescription records were from general practitioner (GP) data coded in Read and British National Formulary (BNF) depending on the data supplier. Diagnoses and procedure records were from hospital in-patient records only and coded in ICD-9/10 and OPCS-3/4 respectively.

\subsection{Annotation} 
\label{subsec:dataprep}

Treatment failure (TF) was defined as having a TF event within one year of the very first clopidogrel prescription. Clopidogrel prescriptions were identified using the substance or brand names of clopidogrel or respective read codes, while TF events include ischemic stroke, myocardial infarct, stent thrombosis and recurrent thrombosis or stenting. All of these annotations were carefully done by clinical professions using ICD-9/10 and OPCS-3/4 codes. 

Subjects with events occurring within 7 days of the first prescription were excluded as it was unclear whether those events were associated with clopidogrel. The visit had to be through the emergency room to be valid in order to exclude follow up visits from previous events. From the dataset, we found 9,867 subjects with clopidogrel prescriptions. Among them, we labeled 1,824 patients as TF cases and 6,859 as control cases; 1,184 subjects were excluded due to data inconsistencies or ambiguities.  


\section{Methodologies}
\subsection{Data Processing}


We gathered various codes together per subject and organized them into visits with the same date. As a result, a subject has multiple visits and a visit has multiple codes. Table~\ref{table:demo_statistics} shows the statistics of the codes and the visits.

\begin{table}[!t]
  \centering
  \caption{Statistics of structured EHR data in UKBiobank.}
    \begin{tabular}{c|r}
    \hline
     \# of subjects & 502,527     \\
     \# of subjects with 1+ visits & 465,506 \\
     \# of unique visits & 22,063,417 \\
     \# of unique codes & 31,589 \\
     Average \# of visits per subject (max/med/min) & 58.3~(2805~/~13~/~1) \\
     Average \# of codes per visit (max/med/min) & 2.7~(61~/~2~/~1) \\
    \hline
    \end{tabular}%
  \label{table:demo_statistics}%
\end{table}%

Note that the number of codes per visit is relatively small compared to the number of words per sentence in natural language processing (NLP) applications. Also note that the number of visits per patient is relatively large and their order is evident due to corresponding timestamps. Therefore, we flattened the codes of the visits and concatenated them across all visits so that a patient has a sentence-like sequence of codes that can be fed into NLP models. 



The amount of training data, however, for both pre-training and fine-tuning is small compared to that for natural language applications. This can be difficult to get the full expressive power of large language model. Therefore, we propose to use data augmentation process to generate additional data for training NLP models. 

There are traditional NLP data augmentation methods, such as synonym replacement ~\cite{sennrich-etal-2016-improving} or back translation ~\cite{NIPS2015_250cf8b5}, but they are difficult to be directly applied to EHR codes. Furthermore, some data augmentation strategies that generate fake data are subject to criticism especially in healthcare domain because the generated data does not exist in the real world.

Hence, we propose a data augmentation method that generates multiple sequences by rearranging orders of codes within a visit. We proposed to use multiple codes including procedures and prescription codes as well as diagnosis codes. Considering there exists an inherent order between these types of medical events (diagnosis, procedures, and prescription), we keep the order of the medical types while we permute the codes within the same medical types.

\begin{figure}[t!]
  \centering
	  \includegraphics[width=\linewidth]{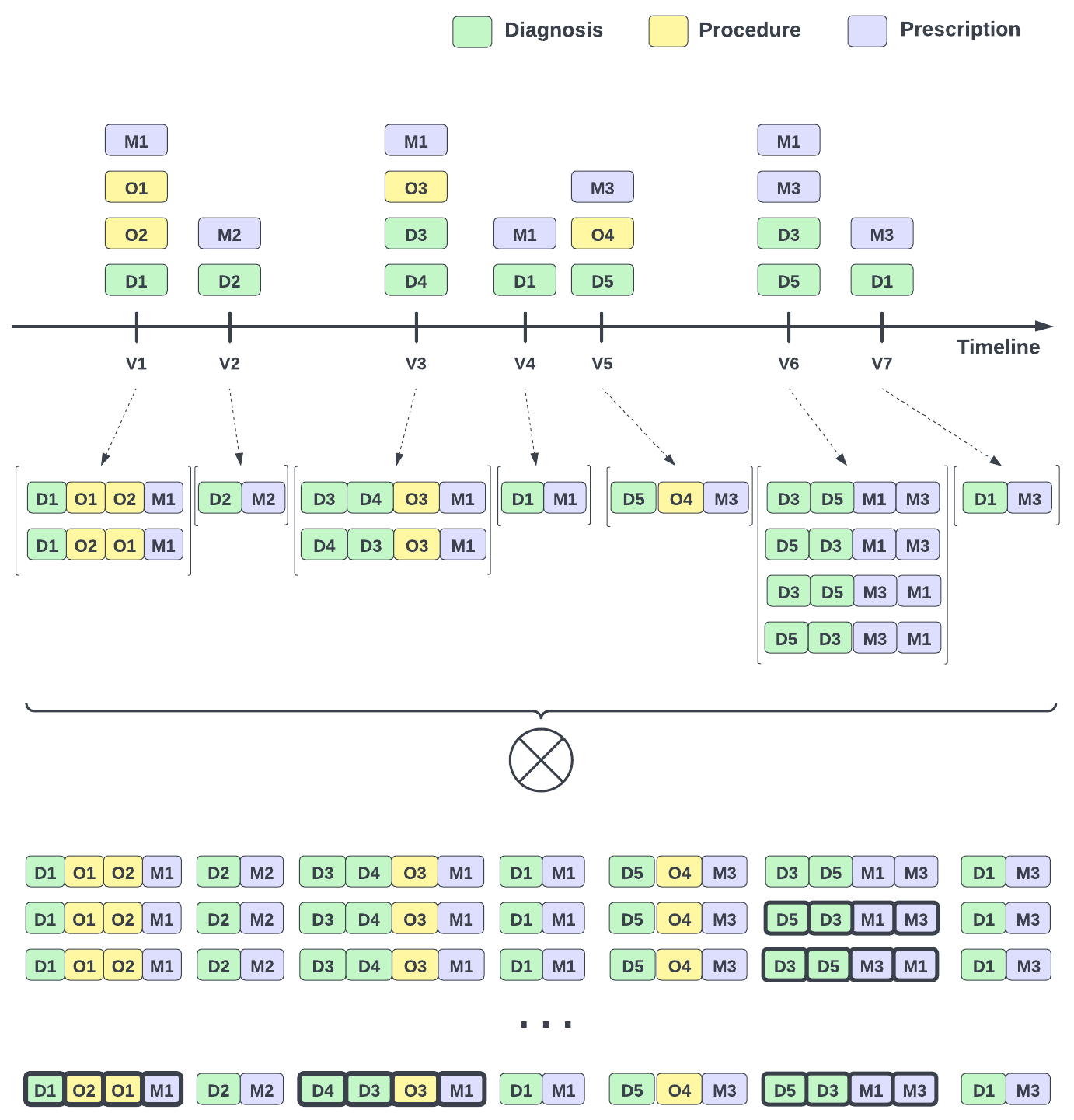}
	  \caption{Proposed data preparation method that augments data. $\bigotimes$ represents a Cartesian product notation. Each box represents medical record, and different colors denote different types of records. Boxes with bold lines simply illustrate different elements from the first sequence.}
\label{fig:ehr_augment}
\end{figure}

Fig.~\ref{fig:ehr_augment} illustrates the proposed augmentation method. It has two steps; the first is to permute medical records within visits, and the second is to combine permuted records to create a sequence. We can formulate out method as follows. Suppose a patient has $n$ visits, $\{V_{1}, V_{2}, \dots, V_{n}\}$ and each visit $V_i$ where $i \in {1, \dots, n}$ is composed of diagnoses, procedures, and prescriptions denoted as $D$, $O$, and $P$, i.e. $V_i = \{D_i, O_i, P_i\}$. When we permute the codes that are within the same types of codes and preserve the inherent order, we can achieve $p_i = |D_i|! \cdot |O_i|! \cdot |P_i|!$ number of codes, where $!$ is the factorial notation and $|\cdot|$ is the number of elements. After code permutation within visits, we combine the permutations to create a sequence, as a result, the total number of possible sequences is $\prod\limits_{i=1}^{n} p_{i}$.

In our experiments, we only selected a subset of all these possible sequences from each patients records for practical reasons because using all the possible sequences is not computationally feasible. We specified the number of the augmented data as $\alpha$, that is, if we choose $\alpha$ as 4, the augmentation method will generate four different sequences of medical records from one patient. In case the number of possible sequences is less than $\alpha$, it only generates available sequences. 

The proposed algorithms can be applied to both of training procedures of BERT, i.e., pre-training and fine-tuning.  

\subsection{Model preparation}
\label{subsec:pretrain}

We applied the proposed data augmentation method to BERT. We used the basic structure of the BERT model architecture consisting of 12 layers of transformer encoders~\cite{devlinBERTPretrainingDeep2019}; it takes 512 tokens as an input and generates a 768-dimensional vector as output.

For pre-training, we used only masked language modeling (MLM) among BERT's pre-training objectives. The next sentence prediction (NSP) which is also widely used in NLP applications is not appropriate in our data preparation setup because we concatenate all the codes of a patient into a single sequence. To build a pre-trained model for BERT, we used the data from the subjects who did not have clopidogrel prescription to learn general population's patterns.

During the fine-tuning, we replaced the last layer of the pre-trained model with a fully connected layer. Since the target task of this study is a binary classification, the last layer is set to have one output node with one node with binary cross-entropy as a loss function. The output of the last transformer encoder layer was fed into a pooling layer, which aggregates the output embeddings into a fixed-length vector representation. In this case, the pooling layer generates a 768-dimensional vector representation of the input sequence.
 

\subsection{Experimental Setup}
\label{experiment}
\begin{figure}[t!]
  \centering	  \includegraphics[width=\linewidth]{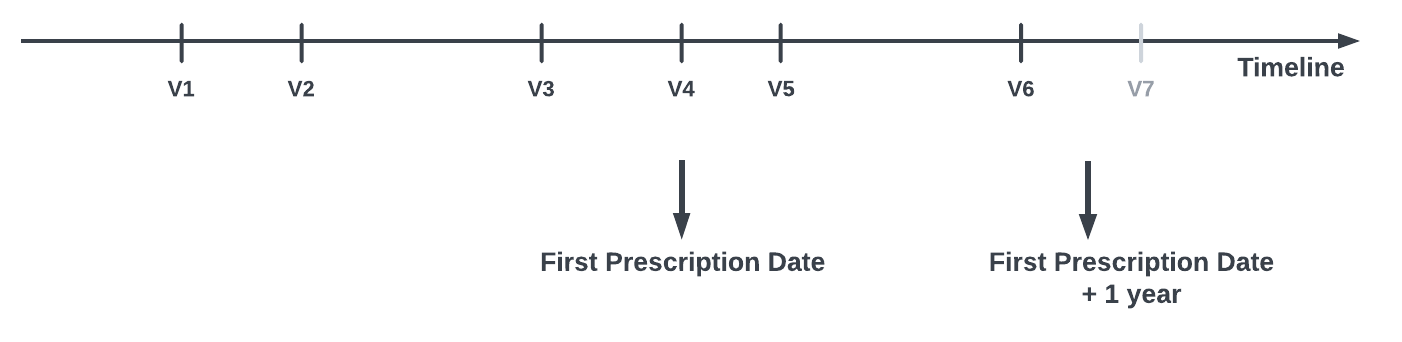}
	  \caption{Detection task; to classify if there exist treatment failures within one year after the first prescription.}
\label{fig:detection_task}
\end{figure}

We have designed an experimental task to detect treatment failures within one year after the first prescription of clopidogrel. The task is to look back from one year after the first prescription to determine if there have been any instances of treatment failure. The model's access to the data is limited to up until one year after the first prescription date. A padding strategy was used to make all the sequences are the same length (512 in this particular case); the most recent codes were used if there were more codes for a patient. Fig.~\ref{fig:detection_task} shows an example of the task. Suppose a patient is first prescribed the clopidogrel at visit 4 (V4), then the takes all the medical records up to 1 year after the first prescription day, i.e. from V1 to V7, and predicts if there exist any treatment failures within one year after the first prescription. 

\begin{figure}[b!]
  \centering	    
  \includegraphics[width=0.7\linewidth]{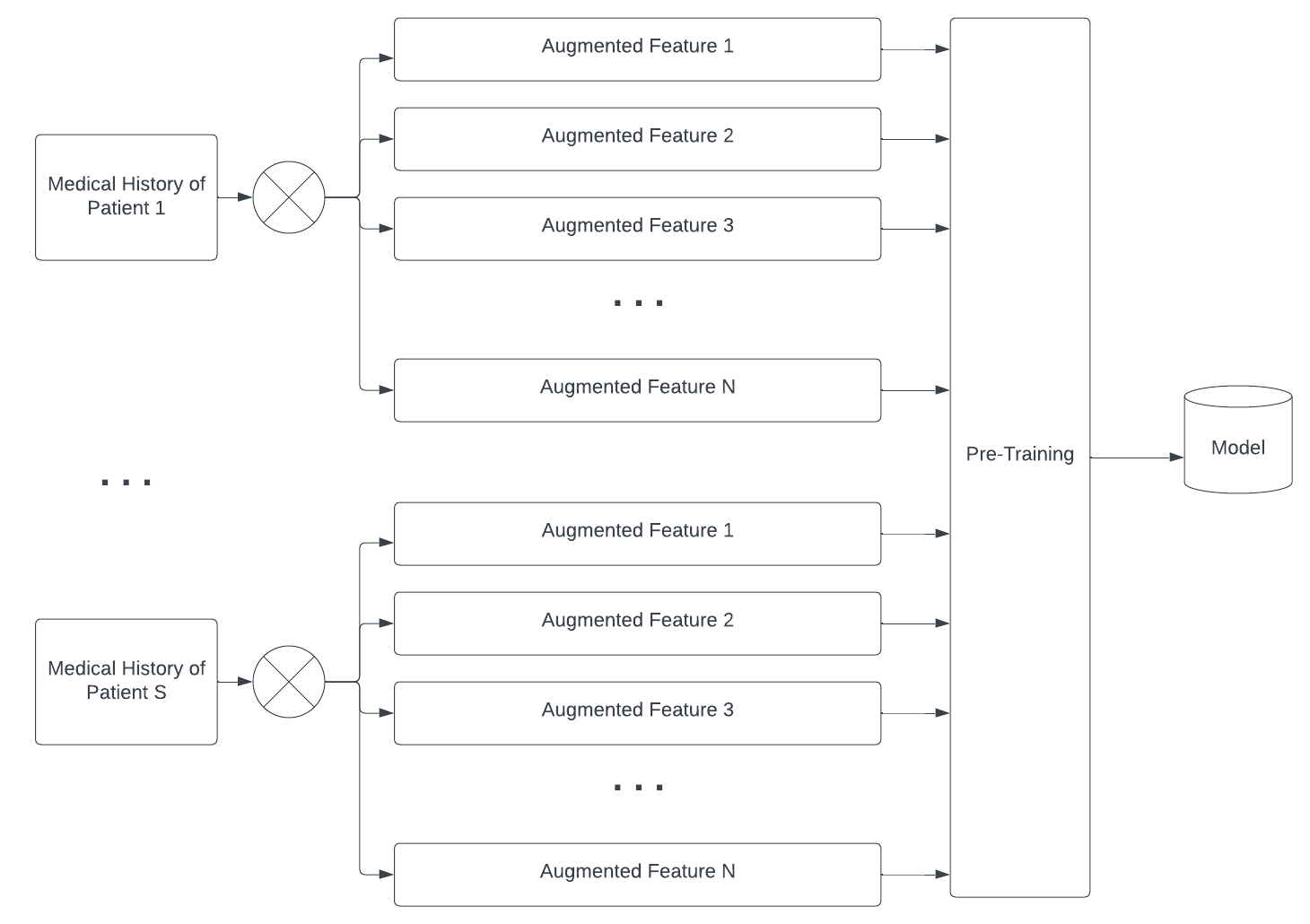}
	  \caption{Pre-training procedure using data augmentation strategy.}
\label{fig:aug_pretraining}
\end{figure}

We pre-trained the model with various augmentation factors. As illustrated in Fig.~\ref{fig:aug_pretraining}, we utilized each of the augmented sequences generated from a particular patient as input for the model. The generated sequences were considered as independent subjects. Since we do not use any label information during the pre-training, we randomly split the data into training and validating data with 80/20 ratio.

\begin{figure}[b!]
    \centering	    
   \includegraphics[width=0.7\linewidth]{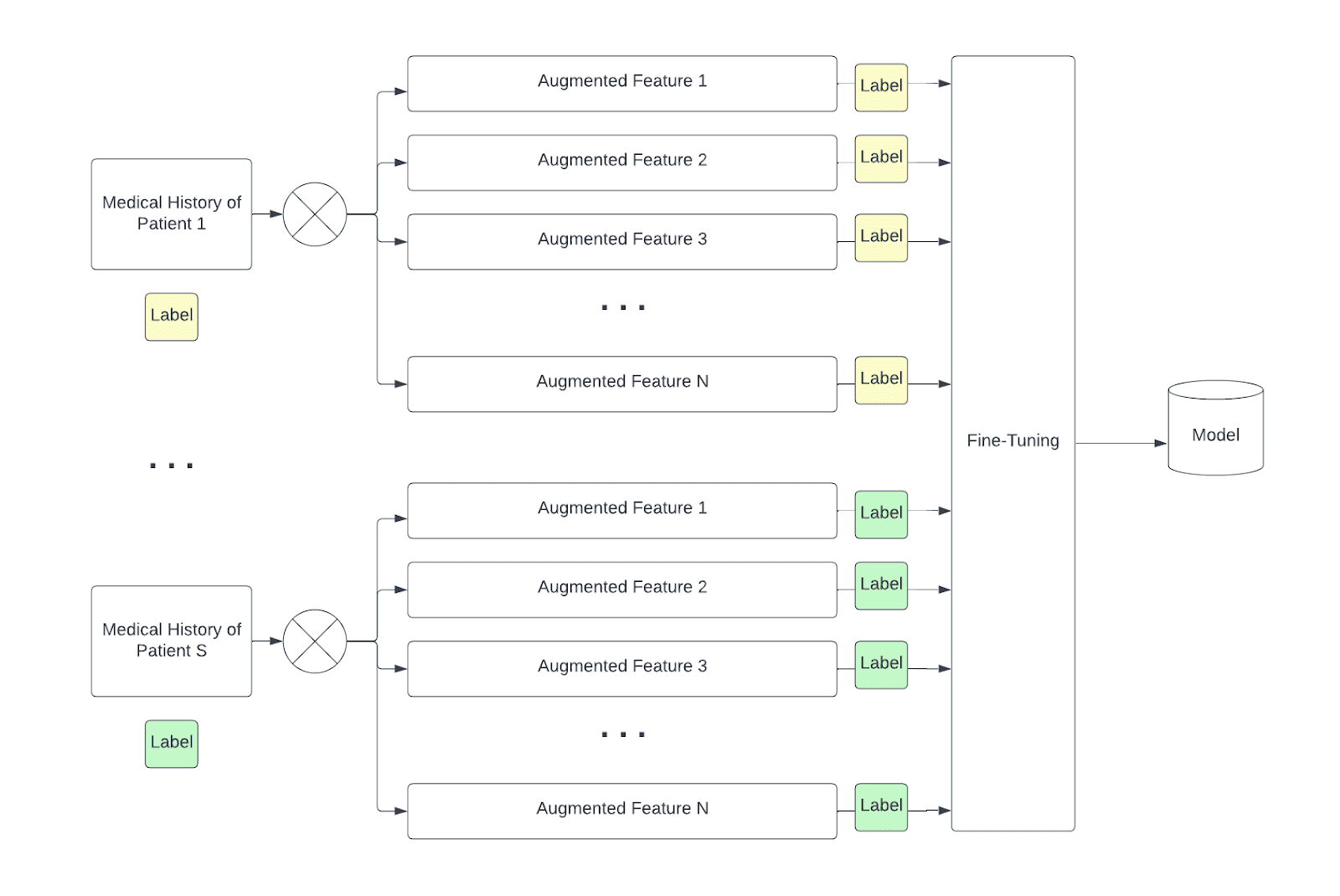}
	  \caption{Fine-tuning procedure using data augmentation strategy..}
\label{fig:aug_finetuning}
\end{figure}

During the fine-tuning procedures, we randomly split the labeled subjects into training and testing data with 80/20 ratio. This split was used across all experiments for fair comparisons. The data augmentation was separately applied in training and testing dataset to avoid augmented data from the same patients present in both datasets. Since the fine-tuning procedure requires label information associated with data, it is reasonable to apply the same label of the patient for the augmented data while treating each sequence as an independent patient. This process is illustrated in Fig.~\ref{fig:aug_finetuning}.

We used receiver operating characteristic (ROC) curves and their area under curve (AUC) to measure the performance of each experiment.

\section{Results and Discussion}
\label{sec:results}

\begin{figure}[b!]
	  \includegraphics[width=\linewidth]{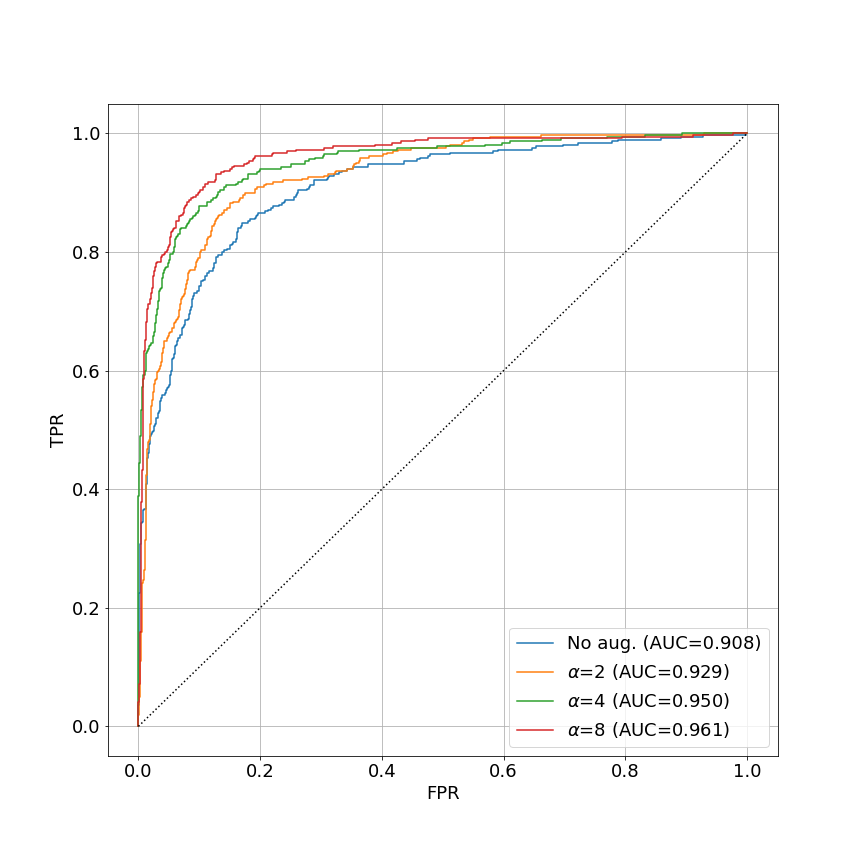}
	  \caption{ROC curves and area under curves based on augmentation factor in pre-training}
\label{fig:pretraining_results}
\end{figure}

\subsection{Impact of augmentation in pre-training}
\label{subsec:aug-pretrain-results}
To examine the performance of augmentation in pre-training, we used the augmentation method only during pre-training but not during fine-tuning. Our experimental results, as shown in Fig.~\ref{fig:pretraining_results}, demonstrate a clear relationship between the amount of augmented data used in pre-training and the model's ability to detect clopidogrel treatment failures. As the augmentation factor increases, the model performance in terms of AUC improves; we were able to achieve 5.3\% absolute improvement compared to the case without augmentation (from 90.8\% without augmentation to 96.1\% with augmentation factor 8). 

We argue that the improvement stems from two folds: more training data by providing increased variety and mitigating ordinal biases imposed by flattening process. This validates our initial hypothesis. The proposed augmentation method allowed the model to learn more robust and generalized representations of longitudinal medical records and leads to significant improvement in a downstream task.


In practice, these findings suggest that data augmentation based on permutation can be a powerful tool for improving the accuracy and reliability of the BERT based model in the healthcare domain, where large and diverse datasets can be difficult to obtain. It also indicates the pre-trained model with augmented data should be delivered as a foundation model~\cite{wornowShakyFoundationsClinical} to be adopted in various downstream tasks for better performance. 

\subsection{Impact of augmentation in fine-tuning}
\label{subsec:aug-finetuning-results}

\begin{figure}[b!]
	  \includegraphics[width=\linewidth]{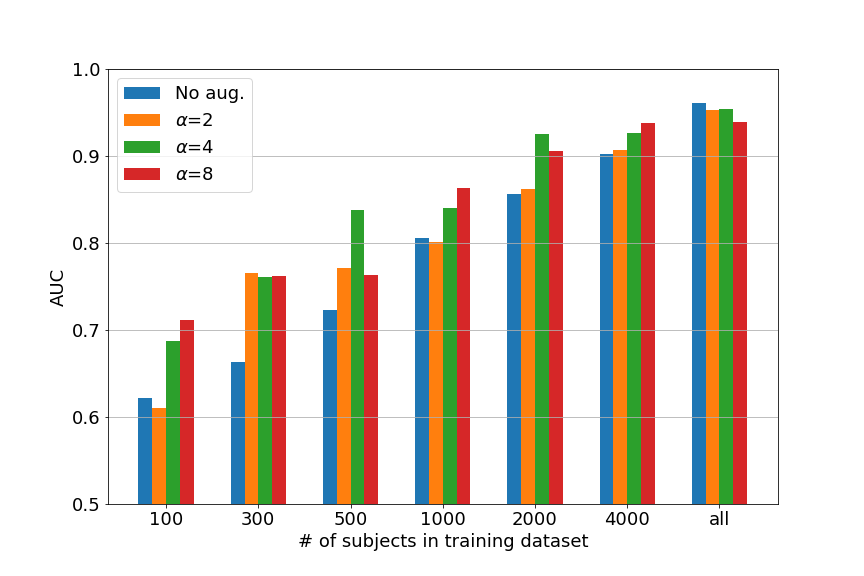}
	  \caption{Performance as the augmentation factor varies with data size}
\label{fig:size_results}
\end{figure}

Subsequently, we investigated the effectiveness of data augmentation in the fine-tuning procedure. Specifically, we performed various experiments in scenarios with limited available data. We used a pre-trained model with an augmentation factor 8, which exhibited the best performance in the previous subsection. 

The corresponding results are shown in Fig.~\ref{fig:size_results}. The results evidently show that performance improves as the amount of available data increases; the best performance could be achieved when all the training dataset is available. More importantly, we also observed that the benefits of augmentation in most of cases. The trend is more pronounced when data samples are scarce with a gradual reduction in performance gain as the size of the data set grows. 
This is particularly noteworthy because collecting sufficient amount of data with appropriate labels can be very challenging. Therefore, based on our experimental results, our approach seems to offer promising outcomes in real-world scenarios.




We also investigated if the data augmentation during the inference procedure would help to improve the detection task performance. This is a test-time augmentation, i.e. we generated augmented test sequences for one test sample and fed them into the fine-tuned model to average scores to represent the test sample. However, the improvement using the test-time augmentation was minimal; we applied the test-time augmentation to all of the models we built, but the maximum improvement we achieved was 0.2\% while there is no performance degradation found.

\section{Conclusions}

The utilization of language models, such as BERT, has become increasingly popular in the medical domain for processing EHR data and related downstream tasks. However, one of the primary challenges associated with leveraging these models is the availability of sufficient pre-training data, which is critical for improving their performance.

To address this challenge, we proposed a novel permutation-based data augmentation approach. Our approach involved generating permutations of input records, where we randomly permute the codes within the same medical types, diagnoses, procedures and prescriptions during a single visit. We hypothesized that this approach would enable the model to learn more robust and generalizable representations of longitudinal medical records by not only providing with more data but also mitigating ordinal biases, thereby improving its overall performance.

Our experimental results demonstrated that the proposed data augmentation approach was effective in improving the performance of BERT-based models. In pre-training, we observed a significant improvement in a downstream application when utilizing the augmented data compared to the case without augmentation. Furthermore, we also found that augmentations can provide performance improvements when there is less data available in the fine-tuning process. We concluded that our method performs well in scenarios where data is scarce, highlighting its potential in practical applications where large amounts of data are difficult to obtain.

Notably, our approach maximizes the use of existing data and avoids generating synthetic data, which is particularly important in the medical domain due to various ethical and legal concerns surrounding the use of patient data. This feature of our approach ensures that the augmented data remains faithful to the original data, minimizing the risk of introducing biases or inconsistencies into the model.

Overall, our study provides valuable insights into the use of data augmentation techniques for improving natural language processing algorithms in medical domains. Our approach has the potential to address the challenges associated with limited data availability, making it a promising solution for practical applications in the healthcare domain. We believe that our findings can inform future research in this area and contribute to the development of more effective and reliable language models for medical applications.

\section{Acknowledgement}
This research has been conducted using the UK Biobank Resource under Application Number 52031.

{\bibliography{mybib}}

\end{document}